\documentclass[sigconf]{acmart}

\AtBeginDocument{%
  \providecommand\BibTeX{{%
    \normalfont B\kern-0.5em{\scshape i\kern-0.25em b}\kern-0.8em\TeX}}}

\usepackage{caption}
\usepackage{subcaption}
\usepackage[ruled]{algorithm2e}

\renewcommand\footnotetextcopyrightpermission[1]{}
\setcopyright{none}
\acmConference[Marble-KDD 21]{Marble-KDD 21}{August 16, 2021}{Singapore}
\begin{document}

\title{Application of Deep Reinforcement Learning to Payment Fraud}


% ------------------Authors------------------

\author{Siddharth Vimal}
\email{Siddharth.Vimal@mastercard.com}

%%\author{Author}

\affiliation{%
  \institution{AI Garage, Mastercard}
  \country{India}
}

\author{Kanishka Kayathwal}
\email{Kanishka.Kayathwal@mastercard.com}

%%\author{Author}
\affiliation{%
  \institution{AI Garage, Mastercard}
 \country{India}
}

\author{Hardik Wadhwa}
\email{Hardik.Wadhwa@mastercard.com}

%%\author{Author}
\affiliation{%
  \institution{AI Garage, Mastercard}
  \country{India}
}

\author{Gaurav Dhama}
\email{Gaurav.Dhama@mastercard.com}

%\author{Author}
\affiliation{%
  \institution{AI Garage, Mastercard}
  \country{India}
}

% ----------------Update Command----------------------
\renewcommand{\shortauthors}{Siddharth Vimal and Kanishka Kayathwal, et al.}

\begin{abstract}

The large variety of digital payment choices available to consumers today has been a key driver of e-commerce transactions in the past decade. Unfortunately, this has also given rise to cybercriminals and fraudsters who are constantly looking for vulnerabilities in these systems by deploying increasingly sophisticated fraud attacks. A typical fraud detection system employs standard supervised learning methods where the focus is on maximizing the fraud recall rate. However, we argue that such a formulation can lead to sub-optimal solutions. The design requirements for these fraud models requires that they are robust to the high-class imbalance in the data, adaptive to changes in fraud patterns, maintain a balance between the fraud rate and the decline rate to maximize revenue, and be amenable to asynchronous feedback since usually there is a significant lag between the transaction and the fraud realization. To achieve this, we formulate fraud detection as a sequential decision-making problem by including the utility maximization within the model in the form of the reward function. The historical decline rate and fraud rate define the state of the system with a binary action space composed of approving or declining the transaction. In this study, we primarily focus on utility maximization and explore different reward functions to this end. The performance of the proposed Reinforcement Learning system has been evaluated for two publicly available fraud datasets using Deep Q-learning and compared with different classifiers. We aim to address the rest of the issues in future work.

%   Fraud is probably the biggest challenge that payment processors face today with cyber-criminals and fraudsters constantly improving the sophistication of fraud attacks. It leads to significant revenue loss and brand damage for different parties in the payment ecosystem including banks and payment processors as well as victimizes the customers through financial losses. Traditionally, fraud detection in payment networks has been formulated as a classification problem with a focus on improving the fraud recall rates of these classification models. A number of papers in the literature have proposed different methods for creating robust features that are immune to model drift using statistical and deep learning techniques. However, this method of problem formulation ignores a number of issues (1) the labels for fraud are usually delayed and usually the financial institution recognizes the fraud much later after the transaction has already been processed (2) the distribution of fraudulent transactions is constantly changing hence the model must be adaptive to these changes (3) the models are not optimized to maximize profit and might end up declining genuine transactions ultimately leading to revenue loss (4) the financial institution loses the label for the transaction once it declines the transaction

\end{abstract}





% \ccsdesc[500]{}
% \ccsdesc[300]{}
% \ccsdesc{}
% \ccsdesc[100]{}



\maketitle
% \usepackage{}

% \acmDOI{}
% \renewcommand\footnotetextcopyrightpermission[1]{}
% \settopmatter{printacmref=false}

\section{Introduction}

With the increasing involvement of businesses and consumers in the digital ecosystem, there has been an exponential rise in digital payments in the past decade fuelled by the variety of payment choices launched by the payments industry. Due to the increasing reach and complexity of the technology involved, fraudsters constantly devise new methods to attack these systems. Many machine learning techniques have already been proposed to tackle this problem, like neural networks \cite{kazemi2017using} and decision trees \cite{varmedja2019credit}, however these techniques can be sensitive to high-class imbalance ratios, changing distributions, and might require re-training. The traditional paradigm of fraud detection solutions in financial institutions consists of formulating it as a classification problem with a focus on improving the fraud recall rates of these classification models. Several papers in the literature have proposed different methods for creating robust features that are immune to model/concept drift using statistical, deep learning, and unsupervised techniques (\cite{lucas2020towards},\cite{carcillo2019combining},\cite{zhang2019hoba},\cite{bahnsen2016feature},\cite{dastidar2020nag}). However, these methods of problem formulation ignore a number of issues: 

\begin{itemize}
    \item \textbf{Utility Maximization} -- the models are not optimized to maximize the utility function and might end up declining genuine transactions, ultimately leading to revenue loss.
    \item \textbf{Non-Stationarity} -- the distribution of fraudulent transactions is constantly changing owing to the emergence of new types of fraud. Also, large-scale fraud events significantly distort these distributions.
    \item \textbf{Asynchronous Feedback} -- the labels for fraud are usually delayed, and the financial institution recognizes the fraud much later after the transaction has already been processed.
    \item \textbf{Counterfactuals} -- the financial institution loses the label for the transaction once it declines the transaction.
\end{itemize}

Moreover, there are practical issues with deploying offline classifier models for fraud since a model trained on historical data might see a loss in performance due to the time required between training and deployment, i.e., the data might become stale.

With the great success that Deep Reinforcement Learning (DRL) methods have achieved in problems with sequential decision making, ranging from achieving professional human-like performance in Atari Games \cite{mnih2015human} to applications in cyber-security \cite{nguyen2019deep} and recommender systems \cite{zheng2018drn}, application of Deep Reinforcement Learning to real-world applications like fraud detection deserves more attention. In this paper, we aim to study the problem of utility maximization in fraud by formulating it as a DRL problem and evaluating different reward functions. The other issues of non-stationarity, asynchronous feedback, and counterfactuals can potentially be solved by various methods available in the DRL literature \cite{foerster2018counterfactual}\cite{walsh2007planning}\cite{igl2020impact}, and it further serves as the motivation to use DRL for fraud detection. We haven't explored these issues in the current study and will consider these in future works. 

However, there are two non-trivial issues that we need to address while attempting this formulation. Firstly, the definition of the reward function must incorporate the utility of money defined for the financial institution deploying the model. Smaller financial institutions with limited budgets might be more risk-averse to potentially fraudulent transactions than larger financial institutions that might prioritize customer experience over fraud costs (especially those offering premium financial products). The famous example which demonstrates the non linearity of the utility of money is the Saint Petersburg Paradox \cite{todhunter1865history}. Various methods in the literature try to estimate this function based on utility elicitation, such as the standard gamble method, time trade-off, and visual analog methods (see chapter 22 of \cite{koller2009probabilistic} for a detailed discussion). However, a detailed comparison of utility estimation methods in the context of fraud detection would probably need a paper of its own hence we will not dwell further into the matter. Moreover, publicly available fraud datasets do not have information that can quantify the utility functions; therefore, we assume that the utility function of the financial institution is risk-neutral, is not dependent upon historically accumulated rewards, and is directly proportional to the revenue earned in the transaction. However, in the real-life scenario, we would need to place a utility distribution over different customer segments (depending on the preferences of the financial institution) and the historically accumulated rewards (depending on the utility of money curve).

% The famous example which demonstrates the nonlinearity of the utility of money is the Saint Petersburg Paradox \cite{todhunter1865history}. There are various methods in the literature which try to estimate this function based on utility elicitation such as the standard gamble method, time trade-off and visual analog methods (see chapter 22 of \cite{koller2009probabilistic} for a detailed discussion). However, a detailed comparison of utility estimation methods in the context of fraud detection would probably need a paper of its own hence we will not dwell further into the matter. Moreover, publically available fraud datasets do not have information that can be used to quantify the utility functions hence we assume that the utility function of the financial institution is risk-neutral, is not dependent upon historically accumulated rewards and is directly proportional to the revenue earned in the transaction. However, in the real-life scenario we would need to place a utility distribution over different customer segments (depending on the preferences of the financial institution) and the historically accumulated rewards (depending on the utility of money curve).

The second issue that we need to address is the definition of state for this problem. While it is intuitive that the historical false decline rate and the fraud rate need to be part of the state, their form of inclusion is not clear. This is because as time passes and we have accumulated a large number of transactions, these metrics will tend to approach a constant value asymptotically, and actions by the agent, even on a large set of transactions, might not significantly alter the state, thus stalling agent learning. Hence it is necessary to introduce some form of decay where older transactions are discarded to calculate false decline rates and fraud rates after some point. We adopt a piece-wise approach where these rates are only calculated only on the current episode and recomputed on the new episode upon completion, and the episode size is a tunable parameter. More details are available in the methodology section.

% The second issue that we need to address is the definition of state for this problem. While it is intuitive that the historical false decline rate and the fraud rate need to be part of the state, their form of inclusion is not clear. This is because as time passes and we have accumulated a large number of transactions these metrics will tend to asymptotically approach a constant value and actions by the agent even on a large set of transactions might not significantly alter the state thus stalling agent learning. Hence it is necessary to introduce some form of decay where older transactions are discarded for the calculation of false decline rates and fraud rates after some point. We adopt a piecewise approach where these rates are only calculated only on the current episode and recomputed on the new episode upon completion and the episode size is a tunable parameter. More details are available in the methodology section.

Based on the factors described above, the contributions of this paper can be summarized as below :
\begin{itemize}
    \item Sensitizing the research community around the problems faced in industrial applications of fraud detection
    \item Formulation of the fraud detection problem as a DRL
    \item A novel tunable reward function that tries to maximize  the revenue and also rewards the agent for a reasonable balance between fraud rate and decline rate
    \item Comparison between the proposed reward function vs. others proposed in the literature along with standard classifiers
\end{itemize}

The rest of this paper is organized as follows. We summarize the
related literature in Section 2, and describe the detailed methodology and architecture in Section 3. We report and discuss
experimental results in Section 4. We talk about future directions and conclude in Section 5.

\section{Related Work}
Supervised machine learning approaches involve building a classification model to identify fraudulent transactions. A number of studies \cite{varmedja2019credit}\cite{mishra2018credit}\cite{ lakshmi2018machine}\cite{ kazemi2017using} compare the performance of multiple machine learning algorithms such as Gradient Boosting, Logistic Regression, Support Vector Machines, XGBoost, Multilayer Perceptron (MLP) on transaction fraud detection task.
\cite{ jurgovsky2018sequence} \cite{roy2018deep} employed Long Short-Term Memory (LSTM) networks to capture the sequential behaviour in fraud detection. Likewise, Convolutional Neural Networks (CNN) based framework is proposed in \cite{fu2016credit} to capture fraud patterns from labeled data. A detailed study presented in \cite{nguyen2020deep} shows that CNN and LSTM perform better than traditional machine learning algorithms in credit card fraud detection \cite{heryadi2017learning} does a similar comparative analysis between CNN, Stacked LSTM, and a hybrid of CNN-LSTM on credit card data of an Indonesian bank. Recently developed attention mechanism is explored in \cite{li2019time}\cite{cheng2020spatio} to detect fraudulent transactions.

Fraud detection can also be formulated as anomaly detection due to the rare occurrence of fraud in the transaction data. Most traditional approaches use distance and density-based methods to detect anomalies such as local outlier factor \cite{breunig2000lof}, isolation forest \cite{liu2012isolation}, K-nearest neighbourhood \cite{angiulli2002fast}. Deep learning is used in anomaly detection by using autoencoders \cite{chen2017outlier}\cite{zhou2017anomaly} or generative adversarial networks \cite{schlegl2017unsupervised} to compute anomaly score using reconstruction error. Some studies \cite{pang2019deep}\cite{ruff2019deep} show improvement in deep anomaly detection with the use of some labeled anomalies in a semi-supervised manner. Few recent papers \cite{pang2020deep}\cite{oh2019sequential} investigate use of reinforcement learning in anomaly detection.

Although there are many deep learning-based methods proposed in fraud detection, the application of reinforcement learning in building fraud systems has not found traction among researchers. \cite{zhinin2020q} uses an autoencoder to learn a latent representation of features and pass them to an agent , which is then trained with a Q-learning algorithm to identify fraud in credit cards. We compare the performance of our reward function with the reward function proposed in this paper. In \cite{el2017fraud}, authors present application of DRL in financial risk analysis and fraud detection while \cite{shen2020deep} proposes alert threshold selection policy in fraud systems using Deep Q-Network.

\section{Methodology}

\subsection{Problem Definition}

Given a dataset \textbf{\textit{D}} = \{$(x_{1}, y_{1}), (x_{2}, y_{2}), (x_{3}, y_{3}), ..., (x_{n}, y_{n}) $\}, where $x_{i}$ is the feature vector for the $i^{th}$ transaction in the dataset and $y_{i}$ represents the corresponding fraud label. Fraud transaction forms the positive class in our datasets i.e $y = 1$ for a fraudulent transaction. We sort the data with respect to time, preserving the sequential aspect and formulate the fraud classification problem as a Sequential Decision Making problem. The agent is given transaction $x_{t}$ at timestep \textit{t}, the agent takes an action of either approving the transaction (\textit{$a_{t} = 0$}) or declining the transaction (\textit{$a_{t} = 1$}). In return the environment provides the agent with a reward based on the current classification performance and the next transaction $x_{t+1}$. The aim of the agent is to be able to classify the transactions such that the utility is maximised such that significant monetary losses due to fraud transactions are avoided and genuine transactions are not declined while also maintaining an optimal balance between the fraud rate (\textit{fr}) and decline rate (\textit{dr}). This is being done using reward function $\mathcal{R}$. Using a Markov Decision Process (MDP) we represent the environment as $\big <\mathcal{S, A, R, T} \big >$ \cite{lin2020deep} \cite{zhinin2020q} with the following definitions:
\begin{itemize}
    \item State $\mathcal{S}$: At time step \textit{t}, the state is $s_{t}$ is the $t^{th}$ transaction $x_{t}$ in the dataset (along with \textit{dr} and \textit{fr} at time \textit{t}). Since, $x_{t}$ contains the attributes of the $t^{th}$ transaction, we will call it the feature vector for $x_{t}$
    
    \item Action $\mathcal{A}$: The action space for this MDP is discrete. We define $\mathcal{A}$ = \{0, 1\}, where the agent can approve ($a_t = 0$) or decline ($a_t = 1$) a transaction.
    
    \item Reward $\mathcal{R}$: A reward \textit{$r_{t}$} is a scalar which measures the goodness of the action $a_{t}$ taken by the agent in the state $s_{t}$. Usually, the reward is positive when the agent takes a preferable action and negative when the action is not desirable. For example, approving a fraud transaction is not preferred, so the agent must be rewarded negatively by the environment. The reward function for the MDP is described in detail in the next subsection.
    
    \item Transition Probability $\mathcal{T}$: The agent takes a decision in the current state $s_{t}$ and is given a new state $s_{t+1}$ by the environment. The new state $s_{t+1}$ is the transaction that occurred just after the transaction $s_{t}$ in the data. We can say that the transition probability is deterministic.
    
    \item Episode: An episode refers to an iteration of the agent interacting with the environment, which includes getting a state, taking action, receiving a reward for the action, and then moving to the next state. An episode ends when the agent reaches a terminal state. For our case, the agent processes transactions one-by-one and reaches a terminal state once it has taken action on \textit{l} (\textit{l} = 500) transactions. This way the agent takes action on $x_{1}, x_{2}, x_{3}, ..., x_{l}$ in the first episode, $x_{l+1}, x_{l+2}, x_{l+3}, ..., x_{2l}$ in the second episode and so on.
    
    \item Decline Rate (\textit{dr}): It is defined as the percent of non-fraud transactions declined during the last \textit{k} transactions processed by the agent.
    
    \item Fraud Rate (\textit{fr}): It is defined as the percent of fraud transactions approved by the agent during the last \textit{k} transactions processed by the agent.
\end{itemize}

\begin{figure*}[h]
\includegraphics[scale=0.7]{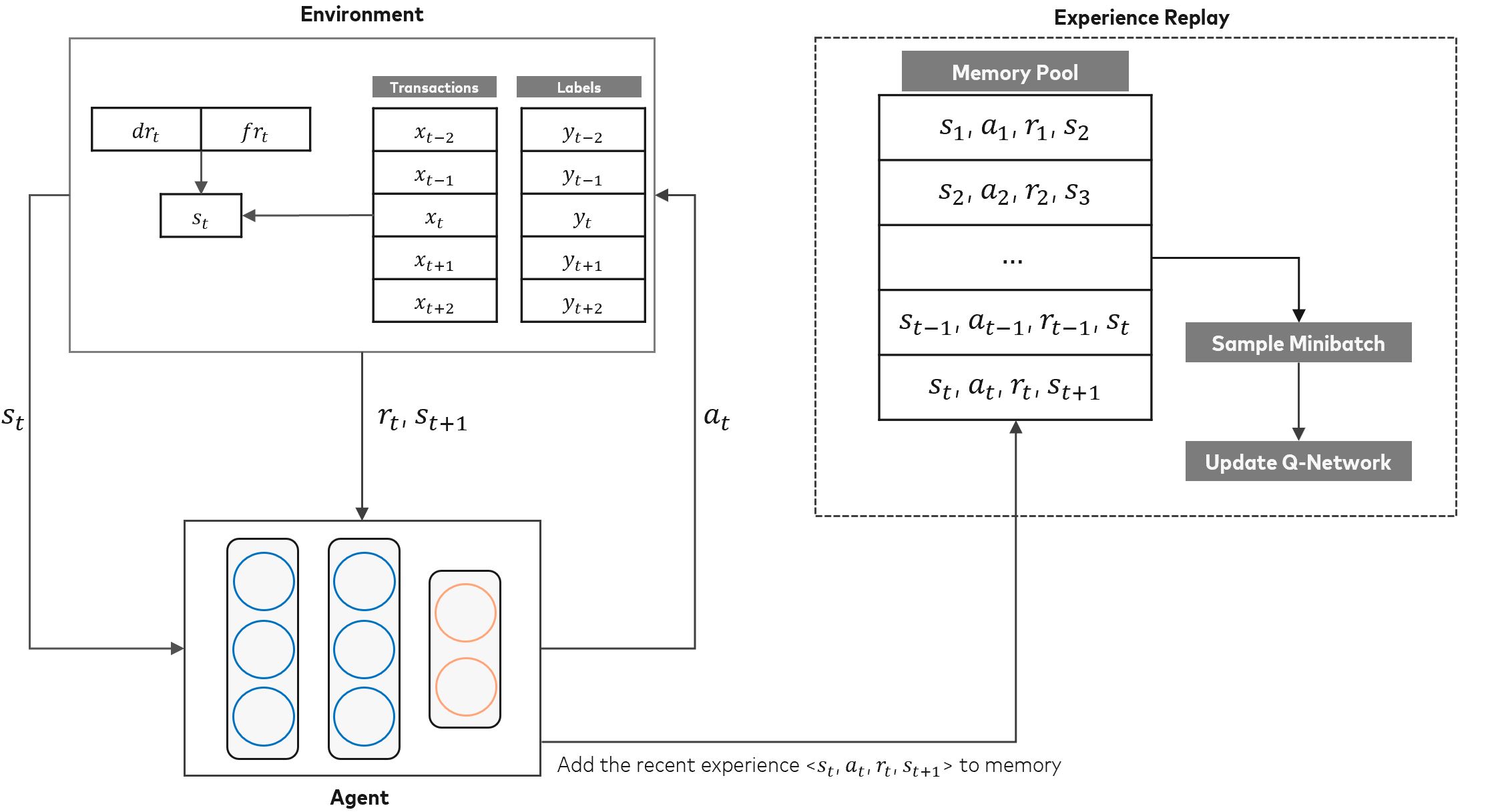}
\caption{Environment design for agent training}
\label{fig:env}
\end{figure*}

The decline rate and the fraud rate are appended to the feature vector of the $t^{th}$ transaction $x_{t}$. This complete vector represents the state $s_{t}$ of our environment at time step \textit{t}. For the experiments, we take \textit{k} = 4000. Figure \ref{fig:env} shows the environment for agent training.

\subsection{Agent} The objective of a reinforcement learning agent is to find an optimal policy $\pi^{*}$ such that it is able to maximise the cumulative rewards $G_{t}$,
\begin{equation}
G_{t} = \sum_{m=0}^{\infty} \gamma^{m} r_{t+m}
\end{equation}
The choice of agent for experiments is the Deep Q-network (DQN) \cite{mnih2015human}, which uses a deep neural network to approximate the optimal action-value function given by: 

\begin{equation}
Q^{*}(s, a) = \max_{\pi}\mathbb{E}[r_{t} + \gamma r_{t+1} + \gamma^{2} r_{t+2} + ...  | s_{t} = s, a_{t} = a, \pi],
\end{equation}

which gives the maximum sum of rewards $r_{t}$ discounted by $\gamma$ at each time step t and following the policy $\pi$ = {P(a|s)}. The neural network learns the parameters $\theta$ by performing Q-learning updates iteratively. At an iteration \textit{i}, the loss function is:

\begin{equation}
L_{i}(\theta_{i}) = \mathbb{E}_{(s,a,r,s')\sim U(M)} \Big [ \Big (r + \gamma \max_{a'} Q(s', a'; \theta_{i}^{-}) - Q(s,a;\theta_{i})\Big)^{2} \Big]
\end{equation}

where $\gamma$ is the discount factor, and $\theta_{i}$ are the parameters of the Q-network at $i^{th}$ iteration and $\theta_{i}^{-}$ are the parameters of the target network, which is used to calculate the target. The target network parameters are set equal to the parameters of the Q-network after every \textit{K} steps. M denotes the memory pool of the agent where it stores its experiences as $m_{t} = (s_{t}, a_{t}, r_{t}, s_{t+1})$ at time t. We chose deep Q-learning because the action space is discrete. The Q-learning updates take place on mini batches (\textit{batch size} = 32) drawn uniformly at random from the memory pool. This process where the agent learns from its past experiences is also termed experience replay. We use a double ended queue of fixed length as the memory pool for the agent. This way, the older samples get discarded, and more recent samples are stored in the memory pool M. The agent is trained using the $\epsilon$-greedy policy. The training process begins with $\epsilon$ = 1 and using decay rate $\delta$ we linearly anneal $\epsilon$ to 0.01.

\subsection{Reward Function}
The agent is rewarded with two rewards $r^{m}_{t}$ and $r^{b}_{t}$ after it takes action $a_{t}$ in state $s_{t}$, which guide the agent to find an optimal policy for maximizing the revenue and also maintaining a balance between fraud rate (\textit{fr}) and decline rate (\textit{dr}). The reward function is defined as:
\begin{equation}
\mathcal{R}^{m}(s_{t}, a_{t}, y_{t}) = 
\begin{cases}
+\alpha*log(amount) & \text{if $a_{t} = 0$ and $y_{t} = 0$}, \\
-1*\alpha*log(amount) & \text{if $a_{t} = 1$ and $y_{t} = 0$}, \\
-1*log(amount) & \text{if $a_{t} = 0$ and $y_{t} = 1$}, \\ 
+log(amount) & \text{if $a_{t} = 1$ and $y_{t} = 1$}
\end{cases}
\end{equation}
The above reward function is inspired from the working of the four party model for payments, where the merchant, acquirer, payment network and the issuer bank involved in a payment keep a small cut of the payment they process. The issuer takes the largest cut (close to 2\% of the transaction amount), but also takes the highest risk. This serves as the base for the \textit{monetary reward} $r^{m}_{t}$. The reward $r^{m}_{t}$ is defined in a manner to be consistent to a banks revenue model on credit/debit cards, where the banks charge an interchange fee to the merchant for every non-fraud transaction approved but suffer a complete loss in case of an approved fraudulent transaction (except for cases of merchant fraud). We keep $\alpha = 0.02$ for our experiments. Second reward $r^{b}_{t}$ is given based on the balance between the fraud rate and the decline rate of the system. This reward is defined as:
\begin{equation}
\mathcal{R}^{b}(dr_{t}, fr_{t}) = \frac{1}{8} * \frac{(1+\beta^{2})*(1-dr_{t}) * (1-fr_{t})}{\beta^{2}*(1-dr_{t}) + (1-fr_{t})}
\end{equation}
where $\beta$ is a hyper-parameter with a preferred value close to 1. The reward function $\mathcal{R}^{b}$ is a scaled down weighted harmonic mean of the two quantities we want the agent to minimize. We scale down the harmonic mean by a factor of 8 so that the maximum possible reward from $\mathcal{R}^{b}$ and $\mathcal{R}^{m}$ is approximately same for each episode. This resulted in a more stable training process.

We further compare the performance of our reward function $\mathcal{R} = \mathcal{R}^{m}(s_{t}, a_{t}, l_{t}) + \mathcal{R}^{b}(dr_{t}, fr_{t})$ with reward functions proposed like $\mathcal{R'}$ \cite{lin2020deep} and $\mathcal{R''}$ \cite{zhinin2020q} for classification using reinforcement learning. The reward function $\mathcal{R'}$ is defined as:
 \begin{equation}
\mathcal{R'}(s_{t}, a_{t}, y_{t}) = 
\begin{cases}
+1 & \text{if $a_{t} = y_{t}$ and $s_{t} \in D_{P}$} \\
-1 & \text{if $a_{t} \ne y_{t}$ and $s_{t} \in D_{P}$} \\ 
+\lambda & \text{if $a_{t} = y_{t}$ and $s_{t} \in D_{N}$} \\ 
-\lambda & \text{if $a_{t} \ne y_{t}$ and $s_{t} \in D_{N}$}
\end{cases}
\end{equation}
For $\mathcal{R'}$, authors suggest $\lambda\ = \rho$, where $\rho$ is the class imbalance ratio given by $\rho = \frac{|D_{P}|}{|D_{N}|}$. $D_{P}$ is the minority (fraud) sample set and $D_{N}$ is the majority (non-fraud) sample set. For the other proposed reward $\mathcal{R''}$ the suggested value for $\lambda$ = 0.1.

\subsection{Training the agent}
We construct an environment according to the MDP defined above. During training, the agent uses  $\epsilon$-greedy policy for selecting its actions. The training stops when the agent has taken action on all the transactions in the training data once. Therefore, the number of training episodes is $\frac{|D_{T}|}{l}$, where $l$ is the length of an episode. The agent training process is given in Figure \ref{training_algo}. The test environment is similar to the train environment, but the agent doesn't use the experience replay in the test environment. Also the actions for test observations are chosen based on the q-values predicted by the Q-network (action with the max q-value is taken by the agent).

\begin{algorithm}
\caption {Training Environment Algorithm}
\label{training_algo}
\KwData{Training Data}
initialize environment variables \textit{fr}, \textit{dr} to 0\;
initialize memory pool \textit{M}\;
initialize parameters $\theta$\;
initialize $s_{1} = (x_{1}, fr, dr)$\;
\For{episode $e=1$ to N}{
    \For{t=1 to l}{
        select $a_{t}$ with $\epsilon$-greedy policy\;
        $r_{t}$, $s_{t+1}$ = step($s_{t}, a_{t}, y_{t}$)\;
        update $fr, dr$ and $s_{t} = s_{t+1}$\;
        store <$s_{t}, a_{t}, r_{t}, s_{t+1}$> to $M$\;
        randomly sample $batch$ from $M$\;
        experience replay on $batch$\;
        % \If{$t = l$}{
        % break\;
% }
}
}
\end{algorithm}

\section{Experimental Results}
\subsection{Datasets}
We have used two open datasets for credit card fraud transactions. First is European card data (ECD) on Kaggle \cite{dal2015calibrating}. This dataset contains 284,807 transactions and 31 features. It contains 492 fraud transactions which make the dataset highly imbalanced (0.172\%). The 28 numerical features in this dataset are the result of PCA transformation. This has been done due to confidentiality and privacy reasons. The data also provides the \textit{time} and \textit{amount} column. The time column is the seconds elapsed from the first transaction in the data, and the amount is the transaction amount.

The second dataset is the IEEE-CIS fraud dataset (IEEE) on Kaggle \footnote{Available at https://www.kaggle.com/c/ieee-fraud-detection}. It contains real-world e-commerce transactions with a variety of numerical and categorical features. There are 590,540 transactions, out of which 20,663 are fraud transactions, so the class imbalance ratio is roughly 3.5\% for this dataset. This dataset also contains the \textit{time} and \textit{amount} columns.

\subsection{Preprocessing}
All the numerical variables are normalized using the \textit{MinMaxScaler}. For the IEEE data, we use only the numerical columns along with a few aggregated features created. We target encoded three categorical variables.

The data is sorted by the time column, the first 70\% of the transactions become the training set. The following 10\% data is used as a validation set for the supervised algorithms, and the last 20\% is used as the test set for evaluation purposes.

\subsection{Evaluation Metrics}
Our evaluation metrics can be categorized broadly into two categories. First, we rely on standard classification metrics such as precision, recall, and F1 score. A high precision, high recall, and high F1 score is the primary step towards model efficacy. In addition, these metrics are helpful in comparative evaluation with other models out there in the literature. Second, in line with our argument that fraud detection should be framed as a utility maximization problem, we evaluate our performance on dollar values of Non-Fraud, Fraud approve and decline numbers. To explain, a financial institution with high Non-Fraud declines might be at the risk of poor customer experience.
On the other hand, an institution with high Fraud approvals is at the risk of losing a massive amount of money. Further, we include two additional metrics: Approval percentage and Fraud(in bps). A low approval percentage means that the institution is declining many genuine transactions to catch frauds, which is unacceptable in a real-world scenario and may lead to reputation loss of the institution. Also, high Fraud(in bps) means the institution is approving many fraud transactions.

% Our evaluation metrics can be categorized broadly into two categories. First, we rely on standard classification matrices such as precision, recall, and F1 score. A high precision, high recall and high F1 score is the primary step towards model efficacy. In addition, these metrics are helpful in comparative evaluation with other models out there in literature. Second, in line with our argument that the fraud detection should be framed as a utility maximization problem, we evaluate our performance on dollar values of Non Fraud, Fraud approve and decline numbers. To explain, a financial institution with high Non Fraud declines might be at the risk of poor customer experience and on the other hand a institution with high Fraud approvals is at the risk of losing huge amount of money. Further, we include two additional metrics : Approval percentage and Fraud(in bps). A low approval percentage means that the institution is declining a large number of genuine transactions to catch frauds, which is unacceptable in a real world scenario and may lead to reputation loss of the institution. Also, high Fraud(in bps) means a large number of fraud transactions are being approved by the institution.

\begin{itemize}
\item App\%: No. of approved transactions per 100 transactions.
\item F(bps): Approved fraud transactions per basis points.
\item $F_{N}$ app: \$ amt of approved genuine (non-fraud) transactions.
\item $F_{N}$ dec: \$ amt of declined genuine (non-fraud) transactions.
\item F app: \$ amt of approved Fraud transactions.
\item F dec: \$ amt of declined Fraud transactions.
    
\end{itemize}

%For evaluating the performance of a fraud model, it is important for the model to have a high precision, high recall and high F1 score. But, it is also important to check business metrics which drastically affect the customers' experience in online payments like approval percentage (by count) and fraud transactions that were approved in bps (basis points). A better model should have a high approval rate and low fraud bps while not compromising the precision and recall scores very much. \textbf{[Explain app\% and fraud bps here??]}

\begin{figure}[h]
\centering
\begin{subfigure}{0.5\textwidth}
\centering
\includegraphics[width=3in]{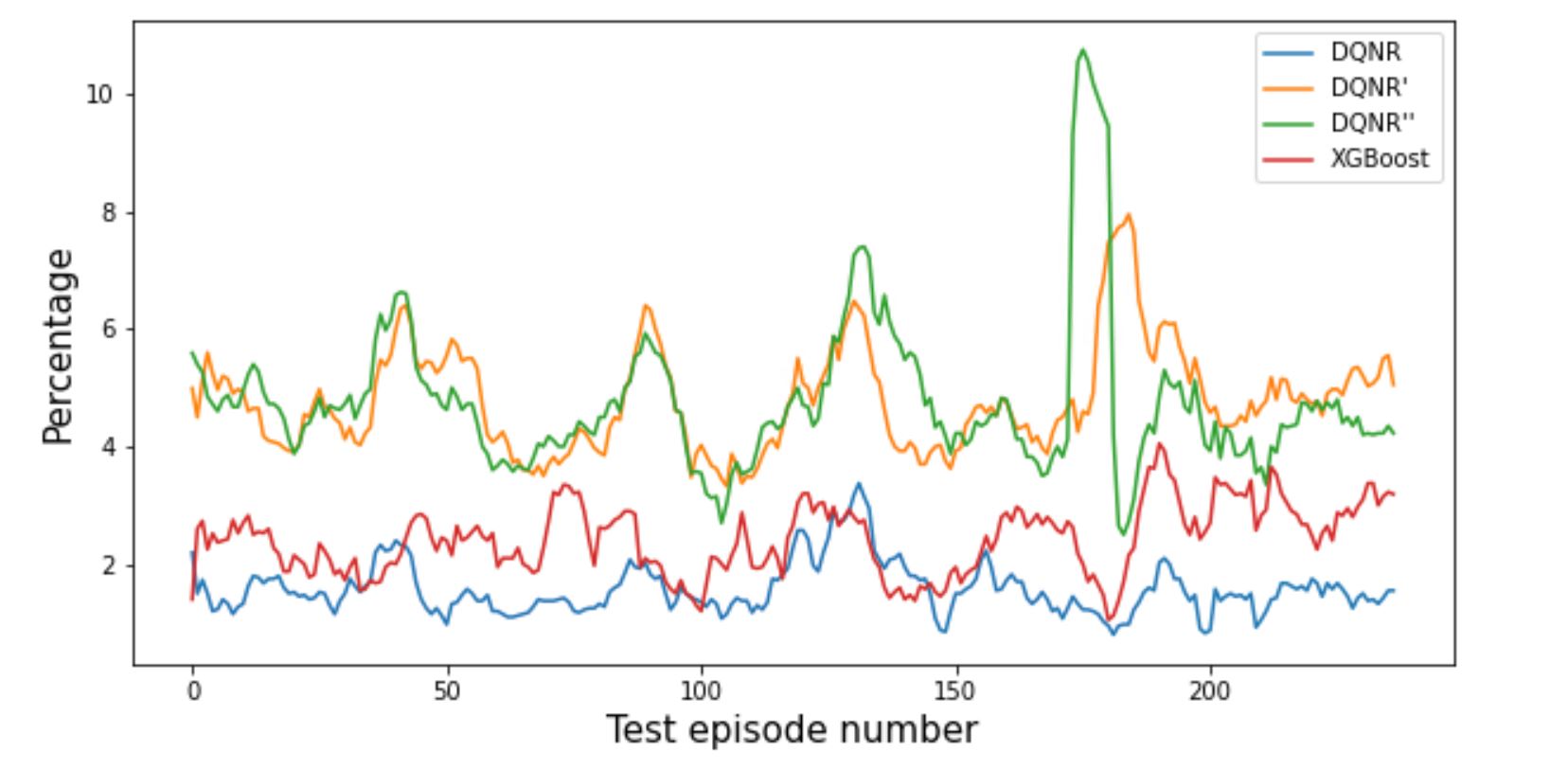}
\caption{Decline Rates ($dr$)}\label{fig:2a}
\end{subfigure}\quad

\begin{subfigure}{0.5\textwidth}
\centering
\includegraphics[width=3in]{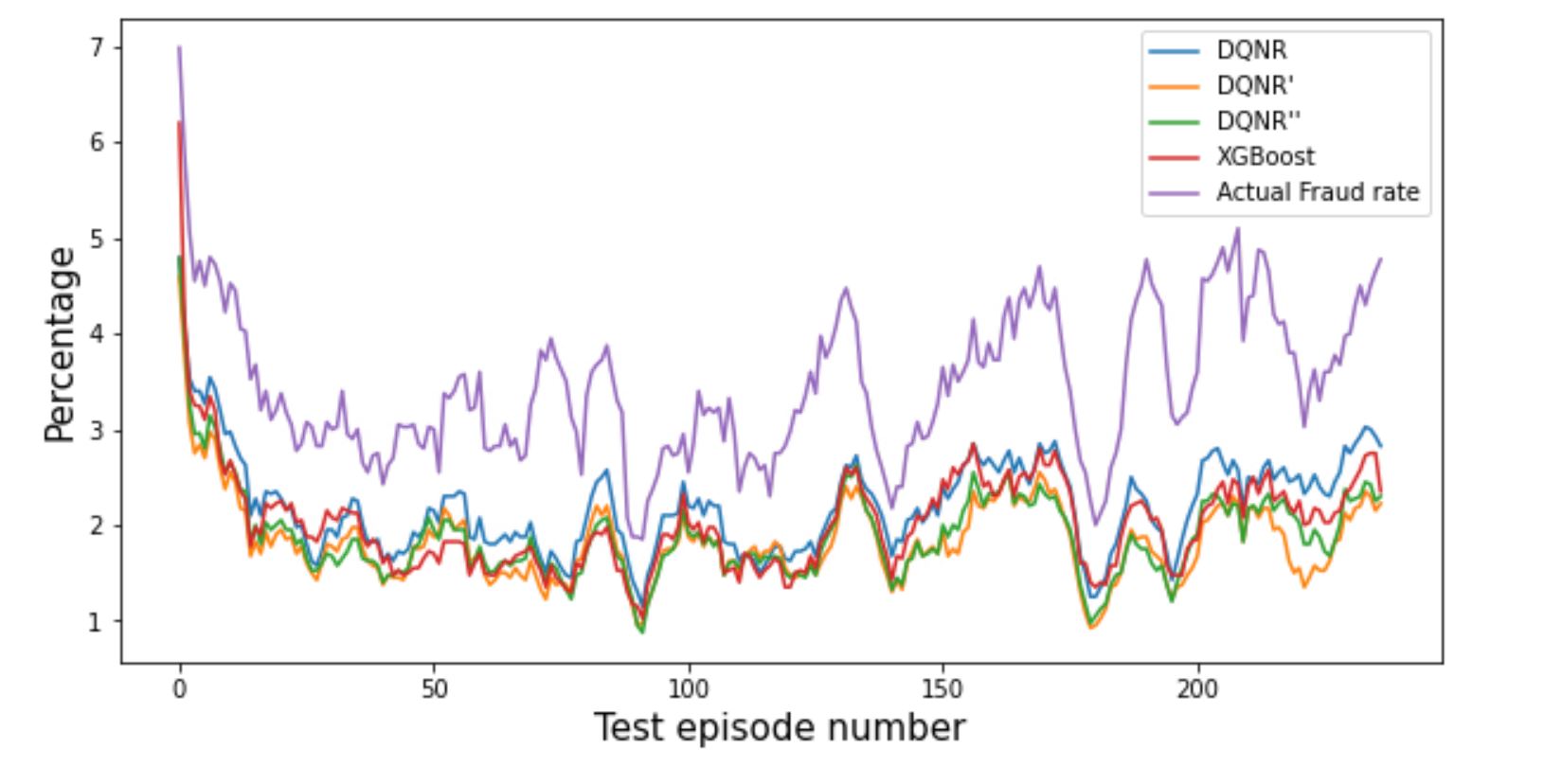}
\caption{Fraud Rates ($fr$)}\label{fig:2b}
\end{subfigure}\quad

% \begin{subfigure}{0.5\textwidth}
% \centering
% \hspace{-0.1in}
% \includegraphics[width=3in]{r_dash_dash_ieee_new.JPG} 
% \caption{DQNR''}\label{fig:2c}
% \end{subfigure}

\caption{Performance of different agents and XGBoost in test environment of IEEE data}
\label{fig:env2}
\end{figure}

% \begin{figure}[h]
% \centering
% \begin{subfigure}{0.5\textwidth}
% \centering
% \includegraphics[width=3in]{r_fr_dr_europe.JPG}
% \caption{DQNR}\label{fig:3a}
% \end{subfigure}\quad

% \begin{subfigure}{0.5\textwidth}
% \centering
% \includegraphics[width=3in]{r_dash_europe.JPG}
% \caption{DQNR'}\label{fig:3b}
% \end{subfigure}\quad

% \begin{subfigure}{0.5\textwidth}
% \centering
% \hspace{-0.1in}
% \includegraphics[width=3in]{r_dash_dash_europe.JPG} 
% \caption{DQNR''}\label{fig:3c}
% \end{subfigure}

% \caption{Performance of different agents in test environment of ECD}
% \label{fig:env3}
% \end{figure}

\subsection{Methods and Performance Evaluation}
To evaluate the performance of a fraud model, we will use precision, recall, and F1 scores. We also compare some business-related metrics to judge the performance of these models.
% \subsection{Parameter Study}
\begin{itemize}
    \item {Agent}: We train three agents $DQNR$, $DQNR'$ and $DQNR''$ with reward functions $\mathcal{R}$, $\mathcal{R'}$, $\mathcal{R''}$ respectively. The Q-network contains two hidden layers with 128 nodes each. For the environment, each episode is of length $l = 500$, the discount rate $\gamma = 0.99$, $\beta = 0.5$. The decay rate $\delta$ is 0.000008 and 0.000004 for ECD and IEEE data respectively. We update the target network every 25 episodes and the length of the memory pool $M$ for the agent is 75000. The Q-network uses huber loss with Adam optimizer, and learning rate 0.005.
    \item {Deep neural networks} (NN): The neural network is similar to the Q-network but is trained using binary cross-entropy loss. The optimizer used is Adam and the learning rate is 0.0002. We use the validation data for the early stopping of the training process.
    \item {CNN} \cite{lecun1998gradient}: We applied 1-D CNN on the data represented as a feature matrix. The network consists of two convolution layers followed by a dense layer.  
    \item {LSTM} \cite{hochreiter1997long}: Transaction sequences are first created using a rolling window with a window size of 30. A 2 layer stacked LSTM is used to capture the sequential information. This information is passed to a dense layer for the final prediction.
    \item {Random forest classifier} (RF) \cite{liaw2002classification}: It is a popular choice for classification problems. We select the best parameters using a randomized grid search.
    \item {XGBoost} \cite{chen2016xgboost}: Parameters for the XGBoost classifier are selected from a randomized grid search.
\end{itemize}

\begin{table}[ht]
  \caption{Experiment Results for ECD}
  \label{tab:results_ecd}
  \begin{tabular}{c|c|c|c|c|c}
    \toprule
    Model & Precision & Recall & F1 Score & App\% & F(bps)\\
    \midrule
    DQNR(ours) & 83\% & \textbf{76\%} & 79\% & 99.88\% & \textbf{3.16} \\
    DQNR' & 98\% & 68\% & 80.3\% & 99.91\% & 4.22 \\
    DQNR'' & \textbf{100\%} & 37\% & 54\% & \textbf{99.95\%} & 8.26 \\
    NN & 98\% & 61\% & 75\% & 99.93\% & 6.15 \\
    CNN & 100\% & 60\% & 75\% & 99.92\% & 5.27 \\
    LSTM & 92\% & 59\% & 72\% & 99.92\% & 5.45 \\
    RF & 80\% & 75\% & 77\% & 99.88\% & 3.34 \\
    XGBoost & 95\% & 72\% & \textbf{81.9}\% & 99.9\% & 3.69 \\
    \bottomrule
  \end{tabular}
\end{table}

\begin{table}[ht]
  \caption{Experiment Results for IEEE data}
  \label{tab:results_ieee}
  \begin{tabular}{c|c|c|c|c|c}
    \toprule
    Model & Precision & Recall & F1 Score & App\% & F(bps) \\
    \midrule
    DQNR & 48\% & 35\% & 41\% & \textbf{97.5}\% & 229 \\
    DQNR' & 32\% & 41\% & 36\% & 95.5\% & 211 \\
    DQNR'' & 23\% & \textbf{47\%} & 31\% & 93.1\% & \textbf{197} \\
    NN & 52\% & 30\% & 38\% & 98\% & 246 \\
    CNN & 43\% & 38\% & 40\% & 97\% & 221 \\
    LSTM & 35\% & 34\% & 35\% & 96.6\% & 234 \\
    RF & 32\% & 40\% & 36\% & 95.8\% & 216 \\
    XGBoost & \textbf{58\%} & 43\% & \textbf{49\%} & \textbf{97.5}\% & 203 \\
    \bottomrule
  \end{tabular}
\end{table}

\begin{table}[ht]
  \caption{Amount approved and declined ECD}
  \label{tab:results_amount_ecd}
  \begin{tabular}{c|c|c|c|c}
    \toprule
    Model & $F_{N}$ app & $F_{N}$ dec & $F$ app & $F$ dec\\
    \midrule
    DQNR & \$4,459,459 & \$1,398 & \textbf{\$2,402} & \textbf{\$5,327}\\
    DQNR' & \$4,435,166 & \$25,691 & \$2,817 & \$4,912\\
    DQNR'' & \textbf{\$4,460,857} & \textbf{\$0} & \$7,286 & \$444\\
    NN & \$4,435,166 & \$25,691 & \$3,871 & \$3,858\\
    CNN & \textbf{\$4,460,857} & \textbf{\$0} & \$4,564 & \$3,166\\
    LSTM & \$4,435,164 & \$25,693 & \$3,620 & \$4,110\\
    RF & \$4,433,783 & \$27,074 & \$2,442 & \$5,288\\
    XGBoost & \$4,435,164 & \$25,693 & \$2,724 & \$5,005\\
    \bottomrule
  \end{tabular}
\end{table}

\begin{table}[ht]
  \caption{Amount approved and declined for IEEE data}
  \label{tab:results_amount_ieee}
  \begin{tabular}{c|c|c|c|c}
    \toprule
    Model & $F_{N}$ app & $F_{N}$ dec & $F$ app & $F$ dec\\
    \midrule
    DQNR & \textbf{\$15,487,021} & \textbf{\$146,196} & \$481,757 & \$128,177\\
    DQNR' & \$14,981,863 & \$651,355 & \$391,945 & \$217,989\\
    DQNR'' & \$14,153,283 & \$1,479,935 & \textbf{\$323,505} & \textbf{\$286,429}\\
    NN & \$15,442,216 & \$191,002 & \$483,072 & \$126,863\\
    CNN & \$15,278,589 & \$354,909 & \$429,033 & \$180,901\\
    LSTM & \$14,984,025 & \$649,473 & \$422,208 & \$187,727\\
    RF & \$15,260,220 & \$372,998 & \$458,605 & \$151,329\\
    XGBoost & \$15,389,617 & \$243,880 & \$385,909 & \$224,026\\
    \bottomrule
  \end{tabular}
 
\end{table}

We use the validation data to adjust the probability thresholds to get the maximum F1 score from the supervised algorithms. We don't require any threshold adjustment for our agent DQNR, and the environment parameters are also fixed for the two datasets. Although a comparison between supervised learning methods and DRL might not make sense from a theoretical perspective because of the different nature of the two-class of methods (supervised learning being an "instructive" algorithm vs. DRL being an "evaluative" algorithm), the purpose is to compare the utility of these methods for the organization using these methods in production. 

\subsection{Discussion}

We report the performance of different algorithms in Table \ref{tab:results_ecd} and Table \ref{tab:results_ieee} on precision, recall, F1 score, approval and fraud rate on ECD and IEEE datasets respectively. The proposed DQNR method performs better/at par with all other models on the F1 score except XGBoost on both datasets. XGBoost outperforms all other methods on the F1 score. This is primarily because XGBoost being an "instructive" process, has access to complete data during training which allows it to learn a better representation of the data compared to a DRL agent trained in an episodic manner. These problems can be potentially be resolved by handling the distribution shift in offline reinforcement learning \cite{kumar2019stabilizing}, using a better curriculum strategy \cite{narvekar2020curriculum} or by solving for the representation learning problem \cite{stooke2020decoupling}.

An inherent requirement of fraud models is to maintain an optimal balance between approval and fraud rates as they directly affect the customer experience and fraud cost, respectively. Large financial institutions generally have a high-risk appetite and they would want to keep high approval rates at the expense of incurring losses due to fraud, as this is directly proportional to better customer experience. For both the datasets, DQNR maintains a balance between Approval and Fraud rate. In addition, it has the lowest F(bps) for the ECD dataset among all other models while maintaining a comparable approval rate with the XGBoost model.

We also  provide results on monetary metrics in Table \ref{tab:results_amount_ecd} and Table \ref{tab:results_amount_ieee} for these models. An optimal combination of Non-Fraud (approvals/declines) and Fraud (approvals/declines) dollar amounts would provide us with the right business metrics to further study each of the algorithms above.

1. In the case of Non-Fraud approvals and declines, in both the datasets, the performance of the DQNR model is better/at par. In the Non-Fraud declines of the IEEE dataset, there is a substantial difference in dollar amounts compared to other models. As Non-Fraud declines are the genuine transactions that affect customer experience, this is a priority for major financial institutions.

2. In case of Fraud approvals and declines, the performance of the DQNR model on ECD beats all other models. For IEEE dataset, for DQNR model, the Fraud approval/decline numbers are not able to beat DQNR' and DQNR".
In the next section, we perform a hyperparameter study to understand how $\beta$ is controlling our approval rate, non-fraud declines, and Fraud declines.

We can also compare the stability of three agents - DQNR, DQNR', DQNR'' and XGBoost (XGBoost evaluated in episodic manner) based on Figure \ref{fig:env2} where we draw the actual fraud rate against the agent fraud (\textit{fr}) and decline rate (\textit{dr}) as we progress in terms of episodes for the test set. We can see DQNR' and DQNR'' behave erratically with high decline rates compared to DQNR and XGBoost on the IEEE dataset. This may result in undesirable performance as the data distribution changes or new types of fraud are encountered by these agents. 

\begin{figure}[h]
\centering
\begin{subfigure}{0.5\textwidth}
\centering
\includegraphics[width=2.8in]{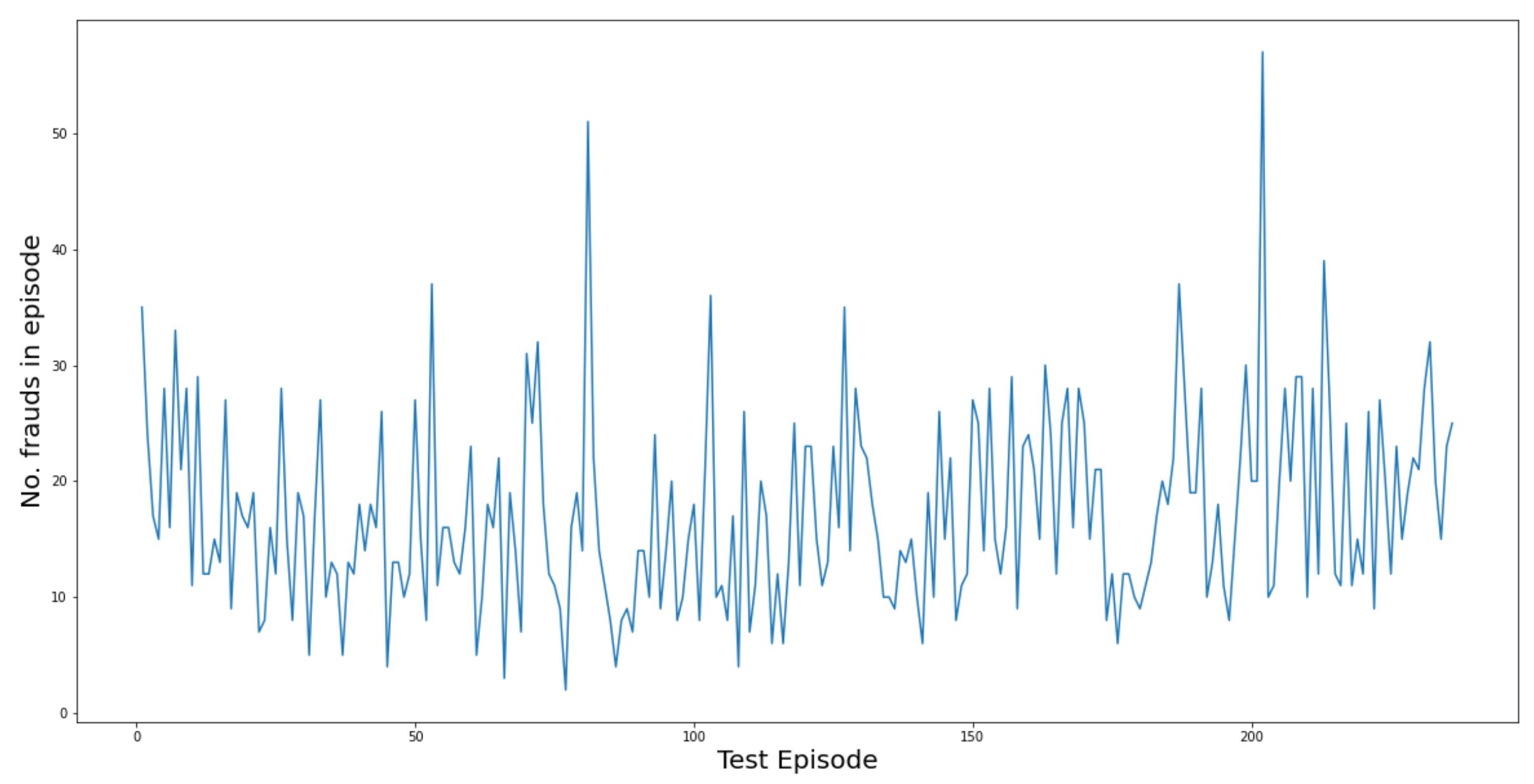}
\caption{IEEE data}\label{fig:4a}
\end{subfigure}\quad

\begin{subfigure}{0.5\textwidth}
\centering
\includegraphics[width=2.8in]{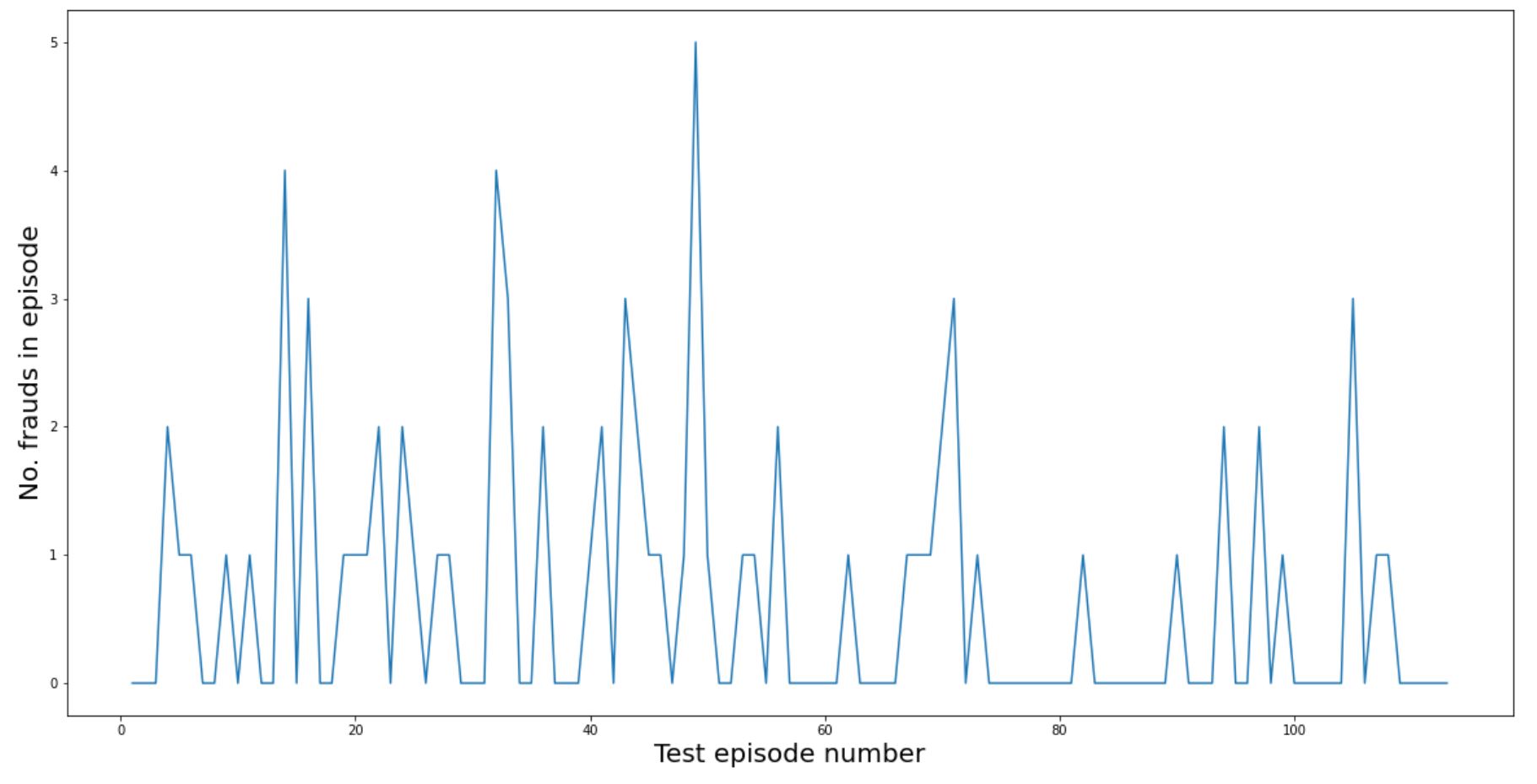}
\caption{ECD}\label{fig:4b}
\end{subfigure}\quad

\caption{No. of fraud examples in test episodes for datasets}
\label{fig:f_no_both}
\end{figure}

Figure \ref{fig:f_no_both} shows how the total no of frauds changes in each episode for the datasets. This type of behavior will be prevalent when these models are trained in an online setting, and our proposed method DQNR has proved to be robust against any such distribution change.

% We can also compare the stability of three agents - DQNR, DQNR' and DQNR'' based on Figure \ref{fig:env2}, \ref{fig:env3} where we draw the actual fraud rate against the agent fraud (\textit{fr}) and decline rate (\textit{dr}) as we progress in terms of episodes for test set. We can clearly see DQNR' and DQNR'' behave erratically with high decline rates in comparison to DQNR on IEEE dataset. This may result in undesirable performance as the data distribution changes or new types of frauds are encountered by these agents. 

% Figure \ref{fig:f_no_both} shows how total no of frauds changes in each episode for the datasets. This type of behavior will be prevalent when these models are trained in an online setting and our proposed method DQNR has proved to be robust against any such distribution change

% 1) What do you see in the results, is one method performing better than the other, what are the potential reasons
% 2) Limitations with our DRL approach like possibly better definition of states, challenges with offline Reinforcement learning (potential overfitting), better representational learning for improving transaction embedding being fed to the agent, limited dataset for offline learning, usage of other DRL methods like actor-critic algorithms

There are certain limitations to using the DRL framework for a fraud detection task. Agents trained on previously collected datasets without any active environment interaction are prone to overfitting as a result of excessive training \cite{singh2020cog}. Their performance is bound by the size of the dataset and highly dependent upon the state and reward definition \cite{wang}.  Further, transaction embedding learned via better representational learning methods \cite{liu2021return} can provide a better state representation and can help the agent to reach the high reward regions of the state space.

% There are certain limitations with using DRL framework for a fraud detection task. Agents trained on previously collected datasets without any active environment interaction are prone to over fitting as a result of excessive training \cite{singh2020cog}. Their performance is bound by the size of the dataset, and highly depend upon the state and reward definition \cite{wang}.  Further, transaction embedding learned via better representational learning methods \cite{liu2021return} can provide a better state representation and can help the agent to reach the high reward regions of the state space.

% \begin{figure}[h]
%     \centering
%     \includegraphics[scale = 0.3]{fno.JPG}
%     \caption{Number of fraud examples in test episodes in IEEE dataset}
%     \label{fig:fno}
% \end{figure}

\subsection{Hyperparameter Analysis}
The proposed reward function has two hyper-parameters $\alpha$ and $\beta$. $\alpha$ is analogous to the cut per transaction that the financial institute gets for each non-fraud transaction it facilitates. $\beta$ is a parameter to adjust the strictness of the agent. Table \ref{tab:beta_res} contains the effects of $\beta$ on the agent's ($DQNR$) performance on IEEE data. This provides a user the flexibility to change the agent's behavior according to their risk appetite. $\beta$ less than one will act loosely for fraud detection and approve most transactions. A higher $\beta$ will be very vigilant for fraud but at the cost of declining genuine transactions.

\begin{table}[ht]
  \caption{Effect on agent DQNR by varying $\beta$ for IEEE data}
  \label{tab:beta_res}
  \begin{tabular}{c|c|c|c|c|c}
    \toprule
    $\beta$ & Precision & Recall & F1 Score & App\% & F(bps) \\
    \midrule
    0.5 & 48\% & 35\% & 41\% & 97.5\% & 229 \\
    1 & 41\% & 38\% & 39\% & 96.8\% & 222 \\
    3 & 35\% & 41\% & 38\% & 96.0\% & 213 \\
    \bottomrule
  \end{tabular}
\end{table}

\section{Future Work and Conclusion}
Fraud detection in payment networks is formulated as a classification problem with a focus on improving the fraud recall rates of these classification models. In this paper, we frame it as a DRL problem and propose a reward function that aims to maximize utility such that significant monetary losses due to fraud transactions are controlled and keeping a check on the decline rate of genuine transactions. We train a RL agent to detect fraud in transaction data while maintaining a balance between the fraud and decline rates. We show that the agent performs well on both the credit card dataset(ECD) and the e-commerce dataset(IEEE) with different class imbalance ratios without the need for aggressive parameter tuning or threshold adjustments. With some modifications, the agent can be used for streaming data and can adapt to changing distributions in a better way. This can solve the issue of re-training fraud models, which is an inherent problem with most classifiers. Furthermore, better algorithms coupled with a more advanced environment and state design might help to improve the performance. The availability of better datasets will also be beneficial for future research.

\bibliographystyle{ACM-Reference-Format}
\bibliography{sample}

% %%
% %% If your work has an appendix, this is the place to put it.
% \appendix









\end{document}